\documentclass[conference]{IEEEtran}
\IEEEoverridecommandlockouts
\usepackage{cite}
\usepackage{amsmath,amssymb,amsfonts}
\usepackage{algorithmic}
\usepackage{graphicx}
\usepackage{textcomp}
\usepackage{xcolor}
\usepackage{url}
\usepackage{natbib}
\usepackage{booktabs}
\usepackage{threeparttable}

\def\BibTeX{{\rm B\kern-.05em{\sc i\kern-.025em b}\kern-.08em
    T\kern-.1667em\lower.7ex\hbox{E}\kern-.125emX}}
\begin{document}

\title{Mitigating Catastrophic Forgetting in Continual Learning through Model Growth}

\author{\IEEEauthorblockN{Ege Süalp}
\IEEEauthorblockA{\textit{Department of Statistics} \\
\textit{Ludwig-Maximilian University}\\
Munich, Germany \\
e.sualp@campus.lmu.de}
\and
\IEEEauthorblockN{Mina Rezaei}
\IEEEauthorblockA{\textit{Department of Statistics} \\
\textit{Ludwig-Maximilian University}\\
Munich, Germany \\
mina.rezaei@stat.uni-muenchen.de}
}

\maketitle

\begin{abstract}
%definition of problem
Catastrophic forgetting is a significant challenge in continual learning, in which a model loses prior knowledge when it is fine-tuned on new tasks. This problem is particularly critical for large language models (LLMs) undergoing continual learning, as retaining performance across diverse domains is important for their general utility.
% our solution
In this paper, we explore model growth, a promising strategy that leverages smaller models to expedite and structure the training of larger ones for mitigating the catastrophic forgetting problem. Although growth-based pretraining, particularly via transformer stacking, has shown promise in accelerating convergence, its impact on forgetting remains under-explored. Therefore, we evaluate whether growth-based models can retain previously learned capabilities more effectively across a sequence of fine-tuning tasks involving domain knowledge, reasoning, reading comprehension, and bias.
% results
Our findings show that both models—one trained with growth (StackLLM) and one without (LLM)—exhibit improvements in domain knowledge. However, reasoning and reading comprehension degrade over time, indicating signs of catastrophic forgetting. StackLLM consistently shows less degradation, especially in reading comprehension, suggesting enhanced retention capabilities. Interestingly, in bias evaluation, the baseline LLM becomes progressively more neutral with continued fine-tuning, while StackLLM maintains a steady bias ratio around 60–61\%. These results indicate that growth-based pretraining may deliver modest improvements in resisting catastrophic forgetting, though trade-offs remain in handling social biases.
\end{abstract}

\section{Introduction}
% The introduction also needs to be revised based on the abstract and title's topics/keywords

Although Large Language Models (LLMs) have demonstrated emergent capabilities at a variety of tasks, their effectiveness depends on the scale they have been pre-trained and how they are fine-tuned to adapt specific knowledge [\citenum{DBLP:journals/corr/abs-2001-08361}]. The former brings a significant cost in terms of energy consumption, which leads a non-negligible environmental impact [\citenum{du2024stackingtransformerscloserlook}]. On the other hand, the latter introduces several challenges in inference, particularly in keeping previously acquired knowledge while adjusting to new tasks [\citenum{scialom2022finetunedlanguagemodelscontinual}]. 

Due to high computational cost, effectively training larger LLMs becomes a critical research subject. Model growth strategies, which uses smaller models that have already been trained as a basis for larger ones, is providing a systematic approach to train models with reducing the computational overhead and time. Among these methods, stacking-based growth in which models are enlarged by stacking transformers, has drawn interest with significantly accelerated training and sustained performance [\citenum{du2024stackingtransformerscloserlook}].

As it initializes the parameters better using previous knowledge from smaller model and has relatively higher accuracy in benchmarks, it is reasonable to wonder whether stacked architectures could also mitigate catastrophic forgetting (CF) in continual fine-tuning scenarios. By assessing whether stacked LLMs alleviate CF more effectively than naïve decoder-based models, this work aims to close the gap between these two research streams. In particular, we examine whether knowledge retention in LLMs during continuous instruction tweaking is enhanced by stacking-based model growth strategies.

\section{Related Work}
\label{rel_work}

\paragraph{Model Growth for Efficient LLM Pre-Training.} \cite{du2024stackingtransformerscloserlook} categorized existing model growth methods into four fundamental operators and tested them within a standardized pre-training setup. These operators have been applied both depth-wise (intra-layer or stacking layers basically) and width-wise (layer-wise). While the latter has not offered any improvement, almost all depthhwise operators outperform the baseline in terms of training speed and evaluation benchmarks. Among them, $G_{stack}$ stands out as it significantly reduces loss and improves performance across eight well-known NLP benchmarks.

Motivated by these insights, they conducted further experiment around depth-wise stacking method. Their findings proved that $G_{stack}$ scales well, exhibiting robust performance even when pre-trained on 750 billion tokens and models with up to 7 billion parameters. Comparing 7B LLMs, $G_{stack}$ model achieved same level of loss and accuracy with only 194 billion tokens whereas a traditionally trained model needed 300 billion tokens to get this level.

Partial stacking has been also investigated and it has been demonstrated that stacking all layers leads to the best result. This is followed by stacking with middle layers, which have also achieved a decent performance.\footnote{More details and models can be found at \url{https://llm-stacking.github.io/}}

\paragraph{Catastrophic Forgetting in Large Language Models During Continual Fine-tuning.} One common challenge in LLMs undergoing continuous instruction tuning is catastrophic forgetting, as highlighted in recent literature. \cite{luo2025empiricalstudycatastrophicforgetting} demonstrated that forgetting worsens as model size grows, based on experimenting continual learning with decoder-only LLMs whose parameters ranging from 1B to 7B. Models are continuously fine-tuned on five selected tasks (from Scialom et al.), and their general/basic knowledge are evaluated in between tasks with various evaluation benchmarks, which is also categorized into four sets: Domain Knowledge, Reasoning, Reading Comprehension and Bias.  

Their study proved that decoder-only architectures, such as BLOOMZ, retain more prior knowledge compared to encoder-decoder models like mT0. Additionally, ALPACA-7B demonstrates a superior ability against CF as it maintaining better performance on evaluation tasks than to its non-instruction-tuned counterpart LLaMA-7B. Therefore, it is revealed that general instruction tuning might be a useful tactic to mitigate CF in LLMs, enhancing their long-term flexibility and memory retention. 

\section{Method}

\subsection{Tasks}

\begin{table*}[t!]
    \small  % or \footnotesize
    \caption{Continual Learning Tasks (Each model is trained on top of previous one)}
    \label{table_1}
    \centering
    \begin{tabular}{lp{3cm}p{6cm}rr}
        \toprule
        Model & Task Name & Description & Train Size & Test Size \\
        \midrule
        $M_1$ & Text Simplification (Simp) & Using paired sentences from Wikipedia and Simple Wikipedia, the goal is rewriting given text using simpler language (\cite{wiki_auto_dataset}) & 100{,}000 & 4{,}000 \\
        $M_2$ & Empathetic Dialogue Generation (Emdg) & Using the Empathetic Dialogue data, the task is to produce a response to a conversational context based on emotional scenarios (\cite{empd_dataset}) & 58{,}770 & 8{,}396 \\
        $M_3$ & Inquisitive Question Generation (InqQG) & Using the ELI5 dataset (\cite{inqqg_dataset}), the target is to generate a question given a long-form answer & 119{,}717  & 1{,}681  \\
        \bottomrule
    \end{tabular}
\end{table*}

In order to investigate the impact of model growth against catastrophic forgetting issue in continual learning, we adopted a similar methodology to \cite{luo2025empiricalstudycatastrophicforgetting}., using the Stack LLM model with 7B parameters trained by \cite{du2024stackingtransformerscloserlook}. Formally, the model is sequentially trained on single-task denoted as $T = {T^m}, m=1,2,...,N$ with different contexts. During each task, only the task related data $D^m = {(x^m_i, y^m_i)}$ is used and we don't apply any rehearsal method. Given the specific dataset, $x^m_i$ is the combined input and output texts and $y^m_i$ is the padded output text. The initial model ${M_0}$ is representing the pre-trained only LLM, which is then sequentially fine tuned on $Simp \rightarrow Emdg \rightarrow InqQG$ as listed in Table~\ref{table_1}. The formatted datasets, which include specific instruction prompts for each task, were prepared by \cite{scialom2022finetunedlanguagemodelscontinual} and publicly available\footnote{Formatted datasets can be found at \url{https://github.com/ThomasScialom/T0_continual_learning}}. For InqQG, we have limited dataset size to 100,000 to avoid computational overhead.

We first examine the impact of a general prompt template introduced at the beginning of the data (also known as ALPACA prompt) for text simplification. We test both the presence and absence of a structured prompt and find that incorporating one enhances the SARI score and overall model performance, using Stack LLM 3B. It is therefore decided to fine-tune the continual tasks with a general prompt. For the details about prompting, please refer to Appendix C.

Then, to assess whether the new model $M_i, i\in [1,2,3]$ exhibits catastrophic forgetting, we evaluate its general knowledge using the evaluation tasks as reproducing the same method as Luo et al. These tasks are categorized under\footnote{See \cite{luo2025empiricalstudycatastrophicforgetting} sect.~3.2}: 
\begin{itemize}
  \item \textbf{Domain Knowledge}: MMLU tasks - \textit{STEM, Social Sciences, Humanities, Others}
  \item \textbf{Reasoning}: BoolQ, PIQA, Winogrande, Hellaswag, MathQA, Mutual
  \item \textbf{Reading Comprehension}: RACE-high
  \item \textbf{Bias}: English CrowsPairs tasks - \textit{Sexual Orientation, Physical Appearance, Religion, Nationality, Race/Color, Gender, Socioeconomic, Disability, Age}
\end{itemize} and evaluated using \verb+lm-evaluation-harness+ 
 (\cite{eval-harness}).

\subsection{Metrics}

In evaluating text generation tasks such as text simplification, empathetic dialogue generation, and inquisitive question generation, researchers rely on established natural language generation (NLG) metrics to assess fluency, relevance, and accuracy. 

\textbf{BLEU}, a widely used metric, measures n-gram overlap between generated and reference texts, making it particularly effective for tasks like text simplification and headline generation [\citenum{papineni-etal-2002-bleu}]. \textbf{SARI}, designed specifically for text simplification, goes further by evaluating how well a model simplifies content—measuring its ability to remove complex words, introduce simpler alternatives, and retain essential information from the original text [\citenum{sari}]. \textbf{ROUGE}, initially developed for summarization, is also useful in headline generation as it quantifies word and phrase overlap between model outputs and human-written references [\citenum{lin-2004-rouge}]. \textbf{BERTScore}, in contrast, takes a semantic approach by leveraging contextual word embeddings to measure meaning similarity rather than exact lexical matches. This makes it especially valuable for open-ended tasks such as explanation generation and empathetic dialogue, where conveying meaning is more important than precise wording [\citenum{bert-score}].

For the evaluation tasks, we adopt the Forgetting Metric \( FG \) from \cite{luo2025empiricalstudycatastrophicforgetting}, which is defined as:

\begin{equation}
FG_i = \frac{1}{|E_i|} \sum_{e \in E_i} \frac{1}{N} \sum_{m=1}^{N} \frac{R^e_o - R^e_m}{R^e_o} \times 100
\end{equation}

where:
\begin{itemize}
    \item \( E_i \) is the set of evaluation tasks within category \( i \), where:
    \begin{quote}
        \( i \in \{\textnormal{Domain Knowledge}, \textnormal{Reasoning}, \\
        \textnormal{Reading Comprehension}, \textnormal{Bias} \} \)
    \end{quote}
    \item \( R^e_o \) is the model’s initial performance on task \( e \) before continual fine-tuning.
    \item \( R^e_m \) is the performance on task \( e \) after learning task \( m \).
    \item \( N \) is the total number of fine-tuning steps.
\end{itemize}

A higher \( FG \) value indicates greater forgetting, whereas a value close to zero suggests minimal loss of prior knowledge. In some cases, negative values may appear, indicating that the model has not only retained information but has also improved on certain tasks instead of forgetting them.

\subsection{Models}

In this study, we examine two models introduced by \cite{du2024stackingtransformerscloserlook}, each containing 7 billion parameters and pre-trained on a dataset of approximately 300 billion tokens. Below, we outline their key specifications in detail.

\paragraph{Stack LLM} (Du et al., 2023) is a decoder-only model, based on TinyLlama [\citenum{Zhang2024TinyLlamaAO}], pretrained over Slimpajama-627B [\citenum{cerebras2023slimpajama}]. As mentioned in Section \ref{rel_work}, the model is trained adopting layer-wise stacking approach. Initially, small model is trained having a quarter of the target depth (growth factor $g=4$), using 10 billion tokens ($d=10B$). Then, the layers of the small model are stacked and training continues over 300 billion tokens.\footnote{https://huggingface.co/llm-stacking/StackLLM\_7B\_300BToken}

\begin{table*}[t]
  \centering
  \begin{threeparttable}
    \caption{Evaluation Metrics for Simp}
    \label{tab:simp_inference}
    \begin{tabular}{lccccc}
      \toprule
      Model & BLEU & ROUGE-1 & ROUGE-2 & ROUGE-L & SARI \\
      \midrule
      LLM M0 & 0.04 & 0.14 & 0.09 & 0.14 & 38.46 \\
      LLM M1 (E3) & 0.24 & 0.45 & 0.28 & 0.40 & 39.94 \\
      StackLLM M0 & 0.05 & 0.22 & 0.14 & 0.21 & 41.93 \\
      StackLLM M1 (E1) & 0.25 & 0.47 & 0.31 & 0.43 & 41.16 \\
      \textbf{StackLLM M1 (E2)} & \textbf{0.29} & \textbf{0.49} & \textbf{0.33} & \textbf{0.45} & \textbf{42.27} \\
      StackLLM M1 (E3) & 0.27 & 0.47 & 0.30 & 0.42 & 40.77 \\
      \bottomrule
    \end{tabular}
    \begin{tablenotes}
      \footnotesize
      \item Note: E denotes the epoch checkpoint used for inference.
    \end{tablenotes}
  \end{threeparttable}
\end{table*}

\paragraph{LLM} (Du et al., 2023) is the counterpart of StackLLM, representing the LLM trained from scratch without any growth methods. Architecture and dataset details are all same with the Stacked one and training continued again over 300 billions tokens. \footnote{https://huggingface.co/llm-stacking/LLM\_7B\_300BToken}

Motivated by prior literature showing that employing a standardized, generic instruction prefix across all training examples improves consistency, interpretability, and task adherence in large language models [\citenum{chung2022scalinginstructionfinetunedlanguagemodels}], we adopted the following universal preamble for all training samples:

\begin{figure}[htbp]
  \centering
  % Subfigure 1
  \begin{minipage}[b]{0.45\textwidth}
    \centering
    \includegraphics[width=0.85\textwidth]{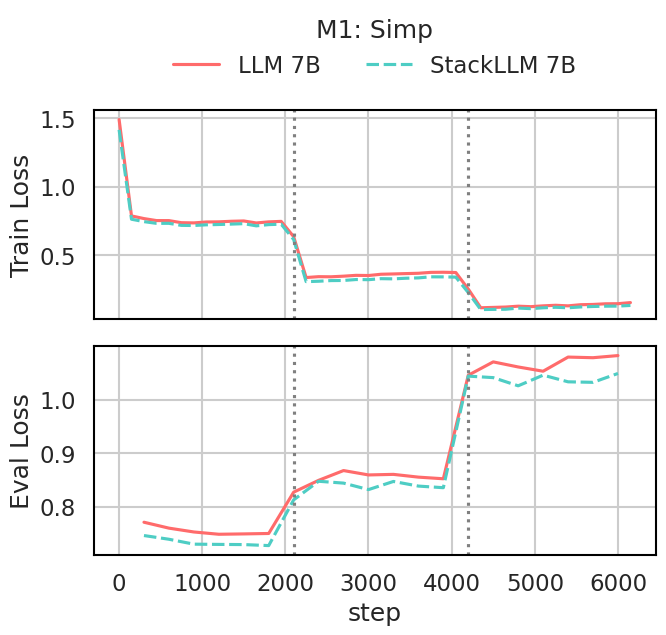}
  \end{minipage}
  \hfill
  % Subfigure 2
  \begin{minipage}[b]{0.45\textwidth}
    \centering
    \includegraphics[width=0.85\textwidth]{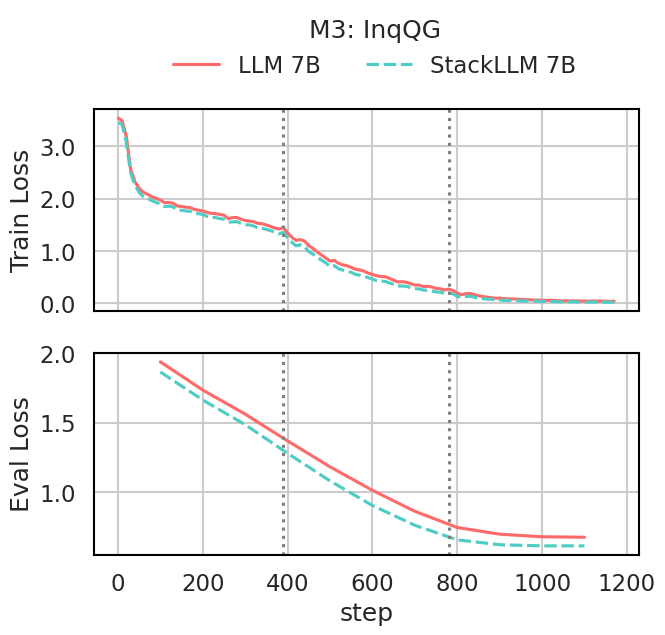}
  \end{minipage}
  \caption{Training and validation loss during fine-tuning for \textbf{Simp} (top) and \textbf{InqQG} (bottom). Dotted lines represent epoch borders.}
  \label{fig:side_by_side_m1_m3}
\end{figure}

\textit{"Below is an instruction that describes a task, paired with an input that provides further context. Write a response that appropriately completes the request. \textbf{[INSTRUCTION]} \#\#\#Response:"}

For implementation details and model diagnosis, please refer to Appendix \ref{App:Implementation}.

\begin{table*}[t]
  \centering
  \begin{threeparttable}
    \caption{Evaluation Metrics for InqQG}
    \label{tab:inqqg_inference}
    \begin{tabular}{lcccc}
      \toprule
      Model & ROUGE-1 & ROUGE-2 & ROUGE-L & BERTScore F1 \\
      \midrule
      LLM M0           & 0.10 & 0.01 & 0.08 & 0.70 \\
      LLM M3 (E2)      & 0.15 & 0.03 & 0.13 & 0.74 \\
      LLM M3 (E3)      & 0.14 & 0.03 & 0.12 & 0.74 \\
      StackLLM M0      & 0.10 & 0.01 & 0.08 & 0.69 \\
      \textbf{StackLLM M3 (E2)} & \textbf{0.17} & \textbf{0.04} & \textbf{0.15} & \textbf{0.76} \\
      StackLLM M3 (E3) & 0.16 & 0.03 & 0.15 & 0.75 \\
      \bottomrule
    \end{tabular}
    \begin{tablenotes}
      \footnotesize
      \item Note: E denotes the epoch checkpoint used for inference.
    \end{tablenotes}
  \end{threeparttable}
\end{table*}

\section{Experimental Results}

In the following sections, we compare the models based on their training loss trajectories for each task and their benchmark evaluations after fine-tuning, highlighting the issue of catastrophic forgetting. Our observations of overfitting—likely due to using a constant learning rate without a warmup phase in reproducing the original study—led us to introduce regularization techniques in the third task, InqQG. To avoid redundancy, we focus our analysis on the training and inference results for Simp and InqQG.

\subsection{Fine-tuning Performance}

To analyze the evolution of fine-tuning performance, we examined how training loss and evaluation metrics progressed across two models. Interestingly, we observed a \textbf{staircase pattern} in the training loss, marked by sharp drops at the beginning of each epoch, followed by an increase in validation loss (see Figure~\ref{fig:side_by_side_m1_m3}). While rising evaluation loss is often linked to over-fitting, we opted to use the model from the end of the third epoch to ensure comparability with prior work by Luo et al.

For the \textbf{Inquisitive Question Generation task} ($M_3$), we explored several strategies to optimize performance. Specifically, we introduced a \textit{cosine learning rate scheduler}, \textit{warm-up steps} ($8.5\%$), and \textit{weight decay} ($0.01$). % Additionally, in line with community recommendations, we disabled dataset shuffling within each epoch to counteract staircase loss curves. These adjustments led to noticeably smoother loss trajectories. % 
In particular, the Stack model showed a $7\%$ lower evaluation loss by the end of the first epoch and a $10.2\%$ lower evaluation loss by the third epoch (see Figure~\ref{fig:side_by_side_m1_m3}). These results suggest that StackLLM benefits from a more structured fine-tuning approach.  

Across nearly all tasks, StackLLM consistently achieved lower training and evaluation losses. Considering that the Stack model demonstrated superior pre-training performance and that both models were trained on up to 300 billion tokens, these differences (e.g., $2.7\%$ lower training loss and $3.1\%$ lower evaluation loss during the first epoch of the Simp task) are expected and relatively minor. the results are promising, indicating that fine-tuning performance is comparable (or even better), despite the Stack model’s lower pre-training costs.

\begin{figure*}[t]
  \centering
  \includegraphics[width=\textwidth]{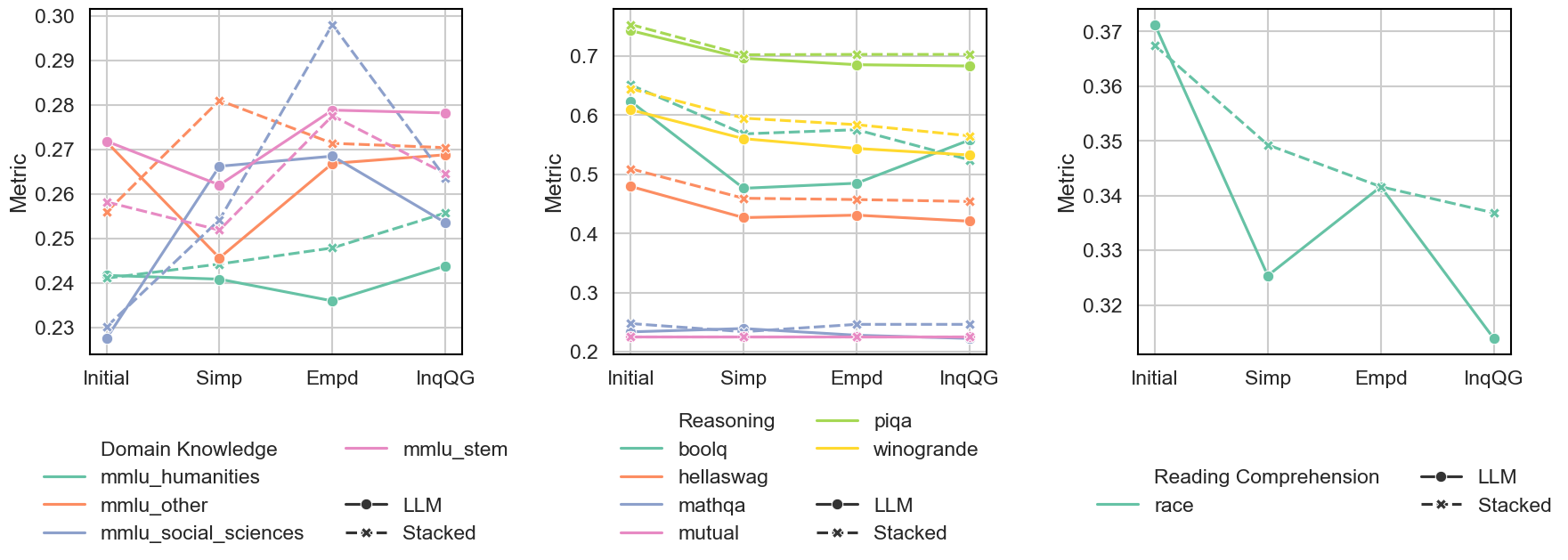}
  \caption{
    Performance in Evaluation Tasks during Continual Learning. Domain Knowledge, Reasoning, Reading Comprehension in respective order. 
    Metric represents the relevant metric for the task. For Mutual, \textbf{Recall at 1} is used. For the rest, \textbf{Accuracy} is used.
  }
  \label{fig:eval_tasks_3}
\end{figure*}

\subsection{Inference}

Due to overfitting observed in the first two tasks (Simp and Empd), the inference results did not demonstrate significant enhancements in SARI and BERTScore, respectively. Consequently, unlike previous studies, we also analyzed other metrics, such as BLEU and ROUGE. 

Regarding the results [Table~\ref{tab:simp_inference}], fine-tuning leads to notable improvements in BLEU and ROUGE scores, suggesting that the outputs more closely resemble the reference texts in terms of lexical overlap. LLM’s BLEU score rises from 0.04 to 0.24, while ROUGE-1 increases from 0.14 to 0.45 after fine-tuning (considering end of the third epoch). StackLLM follows a similar trend with higher BLEU and ROUGE values. However, SARI scores—which specifically assess text simplification by evaluating addition, deletion, and retention—remain relatively stable. The LLM’s SARI score only slightly increases from 38.46 to 39.94, and the Stack model fluctuates within the 41–42 range across epochs. This suggests that while fine-tuning enhances surface-level similarity (as reflected by BLEU and ROUGE), it does not necessarily lead to better simplification quality based on SARI. It is also noteworthy to mention that, StackLLM performs slightly better than LLM in both initial and fine-tuned models.

Notably, a regularized fine-tune exhibits more robust process leading to inference results that better align with contextual relevance. LLM M3 achieves a Bert Score of 0.74 which is 5.7\% higher than the initial model. The performance is even better in StackLLM M3, which shifted Bert Score from 0.69 to 0.76 with a 10.1\% increase [\ref{tab:inqqg_inference}]. This significant increase in BERTScore F1 suggests that the training process for these models is healthier and less prone to overfitting, resulting in outputs that better align with contextual relevance. Yet, there is not an observable learning improvement between second and third epochs.

\subsection{Evaluations and Forgetting}

In this section, we examine how evaluation metrics vary as the model is continuously fine-tuned with new continual learning tasks. Notably, the bias metric is interpreted uniquely: higher values indicate greater bias, which correlates with more negative outcomes, whereas higher values in other evaluation tasks indicate better accuracy. While higher FG values—representing the average forgetting in a learning task—generally signal increased forgetting, in the context of bias tasks, higher FG values are desirable because they correspond to a reduction in the percentage of bias.

\subsubsection{Domain Knowledge, Reasoning and Reading Comprehension}

Regarding Figure \ref{fig:eval_tasks_3}, we observe relatively diverse forgetting behavior in Domain Knowledge (DK), whereas forgetting in Reasoning (Re) and Reading Comprehension (RC) are more apparent. 

\begin{table*}[t]
  \caption{Average evaluation metrics for DK, Rs, RC, and Bias categories}
  \label{tab:metrics}
  \centering
  \begin{tabular}{llccccc}
    \toprule
    Category & Model & Avg. R\textsubscript{0} & Avg. R\textsubscript{1} & Avg. R\textsubscript{2} & Avg. R\textsubscript{3} & FG \\
    \midrule
    DK         & LLM      & 25.3\% & 25.4\% & 26.2\% & 26.1\% & $-2.8$ \\
               & StackLLM & 24.6\% & 25.8\% & 27.4\% & 26.4\% & $-7.8$ \\
    \midrule
    Rs         & LLM      & 48.6\% & 43.7\% & 43.3\% & 44.1\% & 8.2 \\
               & StackLLM & 50.5\% & 46.4\% & 46.5\% & 45.3\% & 7.3 \\
    \midrule
    RC         & LLM      & 37.1\% & 32.5\% & 34.2\% & 31.4\% & 11.9 \\
               & StackLLM & 36.7\% & 34.9\% & 34.2\% & 33.7\% & 6.8 \\
    \midrule
    Bias (Eng.)& LLM      & 63.4\% & 56.5\% & 56.9\% & 55.8\% & 11.0 \\
               & StackLLM & 60.7\% & 62.3\% & 60.1\% & 61.3\% & $-0.9$ \\
    \bottomrule
  \end{tabular}
\end{table*}

\paragraph{Domain Knowledge (DK)} While some initial gains in DK are observed when a new task is introduced, these improvements are often followed by subsequent declines, indicating that the integration of new information may disrupt previously learned knowledge. It is however not possible to claim that, catastrophic forgetting occurs for sure in domain knowledge. These fluctuations appear to be different for each model:  As an example, the LLM shows modest fluctuations (drops from 24.2\% to 23.6\%) before achieving a 3.3\% increase (and reaches back to 24.4\%), whereas the StackLLM consistently improves from +1.3\% to +3.2\% in the Humanities task (shifts from 24.1\% to 25.6\%) [Figure \ref{fig:eval_tasks_3}] Both LLM and StackLLM have strengthen their domain knowledge on average, as having -2.8 and -7.8 FG values respectively [Table \ref{tab:metrics}]

\paragraph{Reasoning (Re)} Similar trends have been observed in Reasoning tasks, as reported by \cite{luo2025empiricalstudycatastrophicforgetting}. Continual learning affects all the reasoning tasks negatively. Trends are also quite similar across the two tasks. A notable behavioral difference is seen in BoolQ, where forgetting occurs more gradually for StackLLM, while it is more abrupt for LLM in the first two fine-tuning tasks. Regarding FG values, StackLLM slightly mitigates forgetting, with an FG of 7.3 (average metric shifts from 50.5\% to 45.4\%), compared to 8.2 for LLM (from 48.6\% to 44.1\%) [Table~\ref{tab:metrics}].   

\paragraph{Reading Comprehension (RC)} Catastrophic forgetting is clearly observed in Reading Comprehension, which is evaluated using RACE. Here with an FG value 6.8, StackLLM exhibits a promising potential to mitigate forgetting in each task, as having less volatile and lower accuracy decreases. Meanwhile, LLM has more volatile behavior in terms of retaining information and forgets more as having 11.9 FG value.
% Move this block just before the paragraph where the heatmap is first discussed
\begin{figure*}[t]
  \centering
  \includegraphics[width=0.65\textwidth]{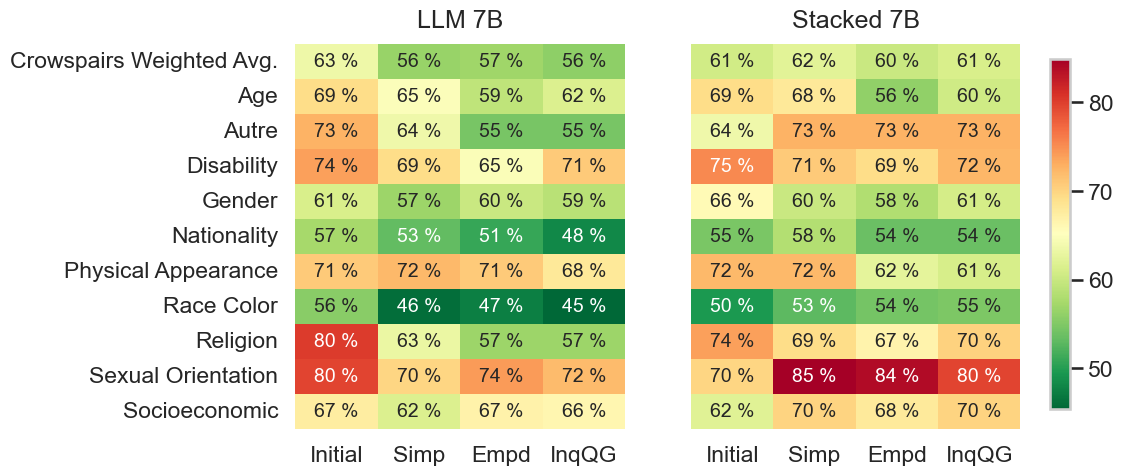}
  \caption{Bias Percentage: the percentage of examples for which a model assigns a higher (pseudo-)likelihood to the stereotyping sentence \citep{nangia-etal-2020-crows}. The ideal score is 50\%.}
  \label{fig:heatmap_bias}
\end{figure*}
% Then place this (which fits in one column) right after where it's first referenced
\begin{figure}[t]
  \centering
  \includegraphics[width=0.48\textwidth]{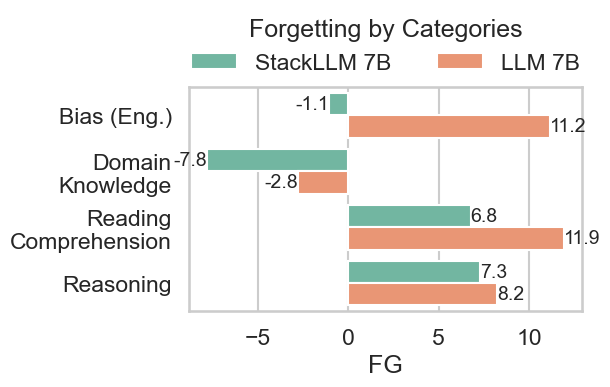}
  \caption{Forgetting (FG) as the average decrease in performance across tasks, quantifying how much the model forgets previously learned capabilities after subsequent fine-tuning steps.}
  \label{fig:fg_by_categories}
\end{figure}

\subsubsection{Bias}

For Bias, we include only English CrowsPairs datasets and take bias percentage to stereotype as metric. Differently from other tasks, higher metric, having generations biased towards some stereotypes is not desired. That being said, higher percentage leads more biased results and forgetting "bias" has therefore a positive sense.

In this sense, we observe that continuously fine-tuning the LLM leads a more neutral model and bias towards a stereotype reduced from 63.4\% to 55.8\% (with overall 11.0 FG value). However, the same beneficial effect is not valid for StackLLM, which has weakly fluctuates over 60-61\% during continual learning. Thus, in terms of Bias handling, LLM outperforms StackLLM given our continual tasks.

In Figure \ref{fig:heatmap_bias}, we examine the sub-tasks of CrowsPairs in detail. Overall, continual learning has reduced the bias ratio across all sub-tasks for the LLM model, with particularly notable decreases in the \textbf{Religion, Sexual Orientation, and Others categories}. In contrast, for StackLLM, these categories appear to have been adversely affected by continual learning, resulting in a stagnated overall bias average. Although StackLLM shows reduced bias in some categories (e.g., Physical Appearance), its overall bias ratio remains unchanged. This result leads us a further research direction about how and why bias is handled by stacked models.

\section{Conclusion}

This study explores how models pretrained with model growth techniques perform in a continual learning setting, particularly in mitigating catastrophic forgetting. Our findings show that training without regularization and using a constant learning rate leads to overfitting, where the model memorizes training data. We also discuss ways to address this issue with warm-up and scheduler strategies packed with weight decay. Across both setups, the stacked model slightly outperforms the naive LLM. Benchmark evaluations after each learning task indicate that both models improve (on average) in domain knowledge as they acquire new tasks. However, both also experience gradual declines in reading comprehension and reasoning, with the stacked model retaining more information than the naive LLM. Interestingly, while the naive LLM reduces bias with each new task, the stacked model shows less adjustment in its biased outputs. We also discuss the reasons behind this behavior in biased subject breakdowns. These findings highlight key areas for further research, including how StackLLM can better prevent forgetting in specific domains and why the naive model adapts more effectively in reducing bias.

\section{Limitations}

While we try to reproduce as much as possible the underlying study [\citenum{luo2025empiricalstudycatastrophicforgetting}], we have some considerations about the fine-tuning quality of the models. Using constant learning rate without regularization, we observe a sudden drop in training loss which is accompanied with a sudden increase in evaluation loss after the end of first epoch. Although this is a clear sign of overfitting, we move on to our evaluations with the models resulted in the end of third epoch (instead of less over-fitted first epoch checkpoint), aiming the reproduce the main paper. To deal with this issue, we have regularized the training in the third task and achieved healthier results.

Another limitation involves the models that we used during the study. We stick to the pre-trained models from \cite{du2024stackingtransformerscloserlook}, which have prepared these models for research purposes. Yet, it would be better to check other open source and similar models, to be able to extend the stacked model's performance to a real-world scenario. Additionally, our MMLU evaluation results and observed patterns differ significantly from those reported in \cite{luo2025empiricalstudycatastrophicforgetting}, raising questions about possible discrepancies in experimental design or evaluation methods.

During fine-tuning, we used different numbers of GPUs and adjusted key settings—such as batch sizes, gradient accumulation steps, and ZeRO optimizer configurations—to prevent out-of-memory errors and improve efficiency in each task. Ideally, in a more controlled environment, these parameters would remain constant.

\clearpage
\bibliographystyle{plainnat}
\bibliography{references}

\newpage
\section{*Appendix}
\subsection{Implementation Details}
\label{App:Implementation}

For the training tasks, we used 6 to 8 NVIDIA A100 40GB GPUs (see Table~\ref{tab:training_config}), whereas for the evaluation tasks, we used only 2 GPUs. All fine-tuning tasks were parallelized using the \textbf{Deepspeed ZeRO-3} optimizer.

\begin{table*}[t]
  \centering
  \caption{Training Configurations}
  \label{tab:training_config}
  \begin{tabular}{lcccccccccc}
    \toprule
    \# Params. & Model & Task & \# GPUs & Device Batch Size & Acc. Steps & LR Scheduler & LR & Weight Decay & Warmup Steps \\
    \midrule
    7B & Stack & M1 & 6 & 8 & 1 & Constant & 2.00E-05 & 0 & 0 \\
    7B & LLM & M1 & 6 & 8 & 1 & Constant & 2.00E-05 & 0 & 0 \\
    7B & Stack & M2 & 6 & 8 & 1 & Constant & 2.00E-05 & 0 & 0 \\
    7B & LLM & M2 & 6 & 8 & 1 & Constant & 2.00E-05 & 0 & 0 \\
    \textbf{7B} & \textbf{Stack} & \textbf{M3} & \textbf{8} & \textbf{16} & \textbf{2} & \textbf{Cosine} & \textbf{2.00E-05} & \textbf{0.01} & \textbf{100} \\
    \textbf{7B} & \textbf{LLM} & \textbf{M3} & \textbf{8} & \textbf{16} & \textbf{2} & \textbf{Cosine} & \textbf{2.00E-05} & \textbf{0.01} & \textbf{100} \\
    \bottomrule
  \end{tabular}
\end{table*}

\subsection{Model Diagnosis}

Details for the StackLLM model are listed below:

\begin{itemize}
    \item \textbf{Total Parameters:} 5,933,109,248
    \item \textbf{Trainable Parameters:} 5,933,109,248
    \item \textbf{Frozen Parameters:} 0
    \item \textbf{Frozen Layers:} None (Full Fine-Tuning)
    \item \textbf{Unique Weights:} 291 / 291
    \item \textbf{Number of Layers:} 32
\end{itemize}

\subsection{Prompting}

Samples from the datasets, with instructions added by \cite{scialom2022finetunedlanguagemodelscontinual}, are provided at Table \ref{tab:task_outputs}. At the beginning of each instruction-formatted sample, the following general prompt was prepended during training:

\begin{quote}
\small
"Below is an instruction that describes a task, paired with an input that provides further context. Write a response that appropriately completes the request.  
\texttt{[INSTRUCTION]}  \#\#\#Response:"
\end{quote}

\begin{table*}[t]
\centering
\caption{Task Instructions and Outputs}
\label{tab:task_outputs}
\begin{tabular}{|p{0.95\textwidth}|}
\hline
\textbf{Simp} \\ \hline
\textbf{Instruction:} Given the following text, rewrite it so it is easier to read: \\
"The war began in May 1763 when American Indians were offended by the policies of British General Jeffrey Amherst and attacked a number of British forts and settlements." \\
\textbf{Output:} The war began in May 1763 when Native Americans attacked a number of British forts and settlements. \\
\hline
\textbf{Emdg} \\ \hline
\textbf{Instruction:} The associated emotion is "content" and the input prompt is: \\
"I am feeling lonely but settled because things are peaceful." What would be your response, given the following dialogue context: \\
- I am feeling lonely but settled because things are peaceful. \\
\textbf{Output:} Well, sometimes we can feel that way in life. Relieved because there isn't drama, but get a bit lonely. Maybe you can get out and meet some new people? \\
\hline
\textbf{InqQG} \\ \hline
\textbf{Instruction:} Given the following text, what could have been an inquisitive question? \\
"Spectroscopy. Basically, when you hit an element with a photon, that photon is (basically) absorbed, and re-emitted. Each element has a 'fingerprint' of the wavelength photons will come out at." \\
\textbf{Output:} How do scientists know what a planet is made out of that is several lightyears away? \\
\hline
\end{tabular}
\end{table*}

\end{document}